\documentclass[referee, pdflatex, main]{main}

\usepackage{algorithmic}
\usepackage{algorithm}
\usepackage{graphicx}
\usepackage{booktabs}
\usepackage{stfloats}
\usepackage{subcaption}
\usepackage{multirow}
\usepackage{textcomp}
\usepackage{array}
\usepackage{colortbl}
\usepackage{amssymb}
\usepackage{amsmath}
\usepackage{xcolor}
\usepackage{soul}
\usepackage{rotating}
\newcommand{\tcp}[1]{\texttt{  // #1}}
\usepackage[many]{tcolorbox}
\usepackage{setspace}
\usepackage{lineno}
\usepackage{caption}

\begin{document}

\title{Glycemic-Aware and Architecture-Agnostic Training Framework for Blood Glucose Forecasting in Type 1 Diabetes}

\author*[1,2]{\fnm{Saman} \sur{Khamesian}}\email{s.khamesian@asu.edu}
\author[1,2]{\fnm{Asiful} \sur{Arefeen}}\email{aarefeen@asu.edu}
\author[1]{\fnm{Maria Adela} \sur{Grando}}\email{agrando@asu.edu}
\author[3]{\fnm{Bithika M.} \sur{Thompson}}\email{Thompson.Bithika@mayo.edu}
\author[1]{\fnm{Hassan} \sur{Ghasemzadeh}}\email{Hassan.Ghasemzadeh@asu.edu}

\affil*[1]{\orgdiv{College of Health Solutions}, \orgname{Arizona State University}, \orgaddress{\street{550 N 3rd St}, \city{Phoenix}, \postcode{85004}, \state{AZ}, \country{USA}}}
\affil*[2]{\orgdiv{School of Computing and Augmented Intelligence}, \orgname{Arizona State University}, \orgaddress{\street{699 S Mill Ave}, \city{Tempe}, \postcode{85281}, \state{AZ}, \country{USA}}}
\affil*[3]{\orgdiv{Department of Endocrinology}, \orgname{Mayo Clinic}, \orgaddress{\street{13400 E Shea Blvd}, \city{Scottsdale}, \postcode{85259}, \state{AZ}, \country{USA}}}

\abstract{
Managing Type 1 Diabetes (T1D) demands constant vigilance as individuals strive to regulate their blood glucose levels to avert the dangers of dysglycemia (i.e., hyperglycemia and hypoglycemia). Despite the advent of sophisticated technologies such as automated insulin delivery (AID) systems, achieving optimal glycemic control remains a formidable task. AID systems integrate data from wearable devices including continuous subcutaneous insulin infusion (CSII) pumps and continuous glucose monitors (CGMs), offering promise in reducing variability and improving time-in-range. However, these systems often fail to prevent dysglycemia, partly due to limitations in prediction algorithms that lack the precision to anticipate impending glycemic excursions. This gap highlights the need for more advanced blood glucose forecasting methods. We address this need with \textit{GLIMMER} (\underline{G}lucose \underline{L}evel \underline{I}ndicator \underline{M}odel with \underline{M}odified \underline{E}rror \underline{R}ate), a modular and architecture-agnostic training framework for glucose forecasting. GLIMMER combines structured preprocessing, region-aware loss formulation, and weight optimization using a genetic algorithm to emphasize prediction accuracy in dysglycemic regions. We evaluate GLIMMER on two datasets: the publicly available OhioT1DM dataset and a new dataset (AZT1D) constructed by collecting data from 25 individuals with T1D. Our extensive analyses show that GLIMMER consistently improves glucose forecasting performance over baseline architectures, enhancing RMSE (Root-Mean-Square Error) and MAE (Mean-Absolute-Error) by up to 24.6\% and 29.6\%, respectively. Additionally, GLIMMER achieves a recall of 98.4\% and an F1-score of 86.8\% in predicting dysglycemic events, demonstrating its effectiveness in high-risk regions. Compared to state-of-the-art models with millions of parameters—such as TimesNet (18.7M), BG-BERT (2.1M), and Gluformer (11.2M)—GLIMMER achieves comparable accuracy while using only 10K parameters, demonstrating its efficiency as a lightweight, architecture-agnostic solution for glycemic forecasting.
}


\maketitle

 
\section{Introduction}
\label{sec:introduction}

Type 1 diabetes (T1D) is an autoimmune condition in which insulin-producing beta cells in the pancreas are destroyed, leading to lifelong dependence on exogenous insulin. An estimated 8.4 million people worldwide live with T1D, accounting for roughly 5--10\% of all diabetes cases~\cite{guo2022growing}. The prevalence of T1D varies significantly across populations and regions, highlighting the need for tailored management strategies to support diverse patient needs globally. Managing T1D is challenging due to the need for constant monitoring and precise insulin dosing to maintain blood glucose within a safe range. Poor glucose control can result in serious complications, including cardiovascular disease, neuropathy, retinopathy, and kidney failure~\cite{atkinson2014type, diabetes2005intensive, karvonen2000incidence}.

\begin{figure}[t]
\centerline{\includegraphics[width=0.85\textwidth]{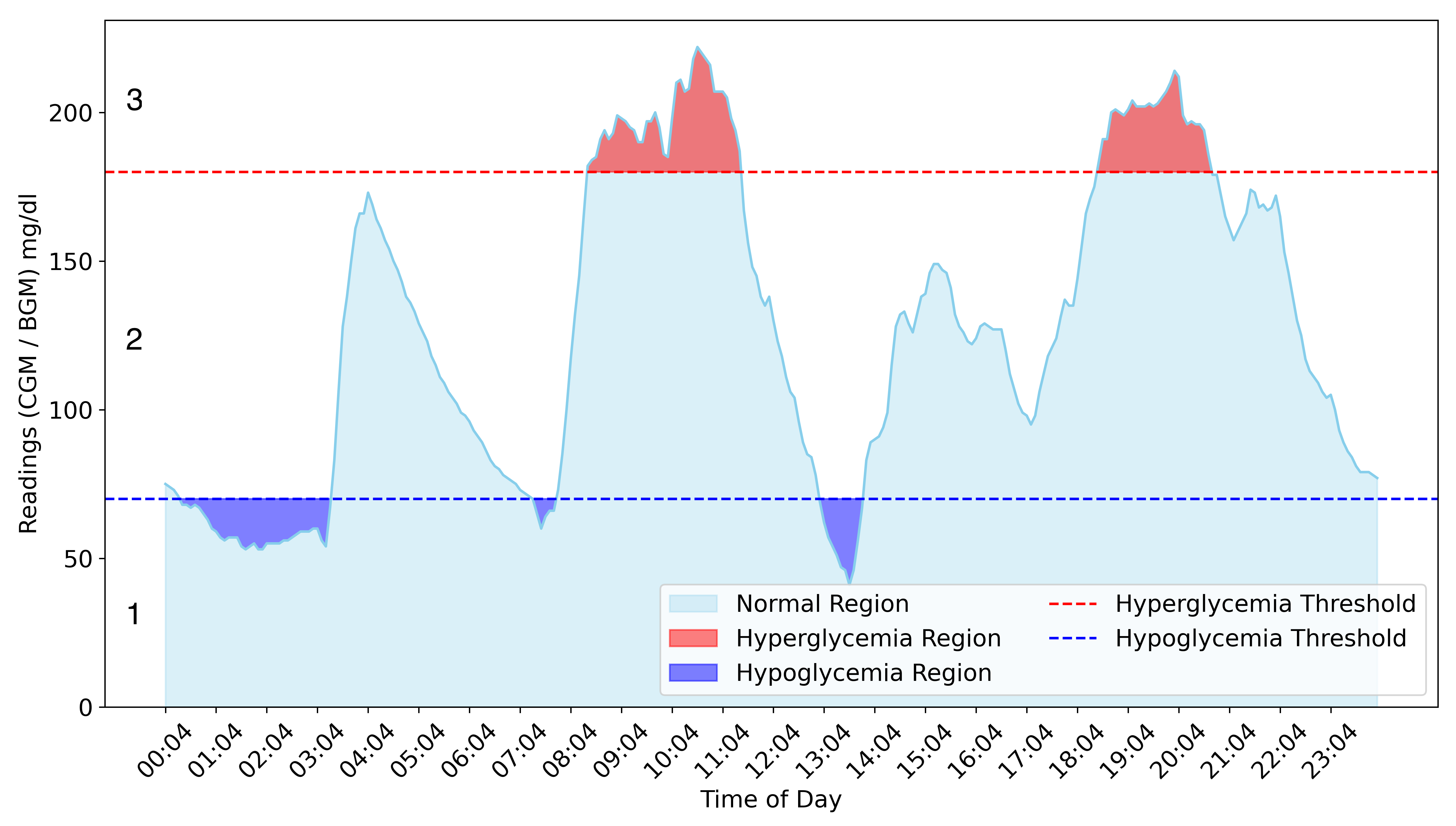}}
\captionsetup{font=small}
\caption{Blood glucose readings captured every 5 minutes. Regions are labeled as follows: (1) Hypoglycemia (below 70 mg/dL, blue), (2) Normal range, and (3) Hyperglycemia (above 180 mg/dL, red). Dashed lines indicate hypoglycemia (blue) and hyperglycemia (red) thresholds.}
\label{fig:1}
\end{figure}

Predicting future blood glucose levels is essential for preventing dangerous dysglycemic events (hypoglycemia and hyperglycemia) and supporting timely, proactive interventions focused on insulin dosing and behavioral modifications (e.g., changes in diet, activity, sleep)~\cite{shroff2023glucoseassist}. It is well established that in AI-powered glucose forecasting models, errors occurring in critical regions—such as misclassifying an impending hyperglycemic event as normal—pose significant health risks~\cite{della2023reducing, nemat2024data, woldaregay2019data, marigliano2024glucose}. In contrast, prediction deviations that remain within the target glucose range (70–180 mg/dL) are generally less consequential~\cite{battelino2019clinical}. Fig.~\ref{fig:1} depicts the clinically defined hypoglycemic, normal, and hyperglycemic regions.

To fully leverage predictive capabilities in real-world settings, these models can be integrated into modern diabetes management technologies. Continuous glucose monitoring (CGM) devices measure interstitial glucose concentrations in real-time, providing frequent data that closely track blood glucose trends and support timely clinical decisions~\cite{desalvo2013continuous}. Automated insulin delivery (AID) systems integrate CGM data with insulin pumps to dynamically regulate insulin dosing using~\cite{limbert2024automated}. Hybrid closed-loop (HCL) systems, a subset of AID technologies, automate basal insulin delivery based on real-time sensor input~\cite{templer2022closed}. When predictive models are embedded within these technologies, they can help overcome the physiological delays of insulin action and glucose sensing, enabling earlier alerts or therapeutic adjustments before glucose levels enter dangerous ranges~\cite{sherr2022automated, renard2023safety}.

Despite advancements in diabetes management technologies, several limitations persist. Existing glucose prediction models often rely on generic architectures and overlook the domain-specific complexities of T1D management. Many models fail to maintain accuracy in clinically significant dysglycemic regions, particularly during rapid fluctuations such as postprandial glucose spikes, where timely intervention is critical \cite{li2019convolutional,aliberti2019multi,sun2018predicting,zhou2021informer,wu2022timesnet}. Although some studies have explored weighted or penalty-based loss formulations to emphasize clinically important regions \cite{wolff2025blood,cichosz2021penalty,de2021integration}, these approaches do not explicitly optimize loss weights to balance predictive accuracy and dysglycemia sensitivity. As summarized in Table~\ref{tab:1}, these limitations highlight the need for domain-specific, clinically-aware designs that prioritize safety-critical objectives. Another key challenge lies in data availability: the high cost and burden of collecting detailed, labeled data limit access to large, high-quality datasets~\cite{del2025availability}. This scarcity creates a cold-start problem for researchers and impedes both model generalizability and personalization~\cite{zhu2022personalized}.

\begin{sidewaystable}[htbp]
\caption{Summary of Related Models in Blood Glucose Forecasting}
\label{tab:1}
\centering
\scriptsize
\renewcommand{\arraystretch}{2.4}
\begin{tabular}{l|p{3in}|p{3in}}
\toprule
\textbf{Model} & \textbf{Key Features} & \textbf{Limitations} \\ 
\hline
FCNN (2022)~\cite{zhu2022personalized} & Simple fully connected network; fast to train and deploy. & Does not model temporal dynamics; limited capacity to capture glucose trends over time. \\ 
CRNN (2019)~\cite{li2019convolutional} & Combines CNN and RNN layers to learn both spatial and temporal features. & General-purpose architecture; lacks focus on clinical events like hypoglycemia and hyperglycemia. \\ 
LSTMs (2019)~\cite{aliberti2019multi,sun2018predicting} & Captures long-term dependencies in time-series; bidirectional versions use future context. & Prioritizes average error; lacks domain-specific loss functions to highlight critical regions. \\ 
MLR (2021)~\cite{zhang2021deep} & Multi-step extension of linear regression; interpretable and fast. & Poor at modeling nonlinear interactions and long-range dynamics; limited scalability. \\ 
Cichosz et al. (2021)~\cite{cichosz2021penalty} & Penalty-weighted loss formulation to emphasize clinically significant glucose regions. & Does not explicitly optimize loss weights for dysglycemia sensitivity. \\
De et al. (2021)~\cite{de2021integration} & Incorporates clinically meaningful metrics into glucose prediction evaluation and model design. & Focuses on clinical integration without systematic loss-weight optimization. \\
Wolff et al. (2025)~\cite{wolff2025blood} & Advocates clinically relevant performance criteria beyond standard accuracy metrics. & Emphasizes evaluation rather than optimized loss design. \\
TimesNet (2022)~\cite{wu2022timesnet} & Learns temporal variations in time-series via 2D decomposition. & High model complexity; lacks explicit clinical interpretability. \\ 
Gluformer (2023)~\cite{sergazinov2023gluformer} & Transformer-based model with uncertainty quantification. & Strong accuracy but still data-hungry and resource-intensive. \\ 
BG-BERT (2024)~\cite{zheng2024predicting} & Self-supervised Transformer for contextual glucose prediction. & Large parameter count; computationally demanding for edge deployment. \\
\bottomrule
\end{tabular}
\end{sidewaystable}

To address these challenges, we propose \textbf{GLIMMER} (Glucose Level Indicator Model with Modified Error Rate), a modular and architecture-agnostic training framework designed for T1D management. Our approach combines structured feature engineering, region-aware loss formulation, and weight optimization to emphasize prediction accuracy in clinically critical blood glucose regions. Specifically, the custom loss function penalizes errors that occur in hyperglycemic and hypoglycemic regions. We formulate the problem of determining the penalty weights for each region (i.e., hyperglycemia, hypoglycemia, and normoglycemia) as an optimization problem over the training loss and employ a genetic algorithm to learn the weight values, placing greater emphasis on errors during dysglycemic events. We demonstrate that this training strategy is architecture-agnostic and can be effectively applied across commonly used forecasting models, including Long Short-Term Memory (LSTM) and Transformer-based networks. To support evaluation beyond existing datasets, we conduct a real-world study and introduce the \textbf{AZT1D} dataset, which comprises 26{,}707 hours of multimodal data from 25 individuals with T1D~\cite{khamesian2025azt1d}. This dataset enables robust evaluation on a distinct cohort and is used in this study to demonstrate the effectiveness of GLIMMER across two independent datasets. Finally, we make our study resources, including the AZT1D dataset~\cite{khamesian2025azt1d} and the GLIMMER source code\footnote{https://github.com/SamanKhamesian/GLIMMER}, publicly available to promote reproducibility.

\section{Related Work}
\label{sec:related_work}

Machine learning and deep learning approaches have significantly advanced blood glucose prediction, enabling proactive interventions and improved glycemic control for patients with T1D. Marigliano et al.~\cite{marigliano2024glucose} demonstrated that integrating predictive alarms with CGM technology reduced hypoglycemic events in adolescents by 40\% and severe hypoglycemia by 60\%, highlighting the clinical benefits of early warnings. Similarly, Vettoretti et al.~\cite{vettoretti2020advanced} showed how AI-based systems can improve outcomes by prompting timely insulin adjustments or dietary changes. In line with these efforts, Arefeen et al.~\cite{arefeen2022forewarning} developed predictive models for postprandial hyperglycemia using controlled feeding trial data.

To improve predictive performance, researchers have explored a range of architectures. Zhu et al.~\cite{zhu2022personalized} applied fully connected neural networks (FCNNs) for their simplicity and speed, while El Khatib et al.~\cite{el2020deep} used convolutional models (CNNs) to extract local patterns in CGM trends. Li et al.~\cite{li2019convolutional} proposed a hybrid CRNN model to capture both spatial and temporal glucose dependencies. Aliberti et al.~\cite{aliberti2019multi} and Lu et al.~\cite{lu2020cnn} demonstrated how LSTM-based models can capture long-term temporal patterns, and Sun et al.~\cite{sun2018predicting} used Bi-LSTM to incorporate future context for improved forecasting. Simpler models such as ARIMA~\cite{plis2014machine} and MLPs~\cite{georga2012multivariate} are still used for interpretability and computational efficiency.

Transformer-based architectures have recently gained attention due to their ability to model long-range dependencies and parallelize computation. Informer~\cite{zhou2021informer} and TimesNet~\cite{wu2022timesnet} have shown strong performance in long-horizon glucose prediction. Gluformer~\cite{sergazinov2023gluformer} enhances this line of work by incorporating uncertainty quantification, while BG-BERT~\cite{zheng2024predicting} applies self-supervised learning to extract contextual information from time-series data. Despite their effectiveness, such models often require large training datasets and substantial computational resources, making them less feasible for deployment on memory- and power-constrained edge devices.

In addition to architectural advancements, several studies have explored loss formulations that incorporate clinical relevance into glucose prediction. Cichosz et al.~\cite{cichosz2021penalty} introduced a penalty-weighted loss function to emphasize clinically critical glucose regions during training. De et al.~\cite{de2021integration} integrated clinically meaningful performance metrics into model evaluation and design, aiming to move beyond conventional accuracy measures. More recently, Wolff et al.~\cite{wolff2025blood} advocated for performance criteria that reflect clinically significant outcomes rather than relying solely on aggregate error metrics such as RMSE. While these efforts represent important steps toward clinically-aware modeling, they do not introduce a systematic framework for explicitly optimizing loss weights to balance overall predictive accuracy and dysglycemia sensitivity.

\section{Methodology}
\label{sec:metholodogy}

\subsection{Model Architecture}
GLIMMER is \textit{architecture-agnostic} and designed to enhance blood glucose forecasting independently of the underlying model architecture. To demonstrate this flexibility, we apply GLIMMER to two representative forecasting architectures: CNN-LSTM and Transformer. These choices reflect both established and state-of-the-art modeling approaches commonly used in time-series glucose prediction.

The CNN-LSTM architecture combines the feature extraction strengths of convolutional layers with the temporal modeling capabilities of LSTM units. This hybrid design effectively captures spatial and sequential patterns within multivariate physiological data. Prior work by Jaloli and Cescon~\cite{jaloli2023long} showed that CNN-LSTM outperforms LSTM, CRNN, and other baselines in predictive accuracy and clinical relevance, especially over longer horizons. In parallel, Transformer-based models have shown strong potential for modeling long-range dependencies via self-attention and have become increasingly popular in time-series forecasting~\cite{zhou2021informer, wu2022timesnet, sergazinov2023gluformer}. By evaluating GLIMMER with both architectures, we highlight its ability to generalize across modeling paradigms while improving performance in critical glycemic regions.

\subsection{Custom Loss Function}
Glucose values are typically divided into three clinically meaningful regions: hypoglycemia, normal, and hyperglycemia as shown in (\ref{eq:glucose_region_class}). Values below the hypoglycemia threshold \(T_{hypo}\) indicate dangerously low glucose levels, while values above the hyperglycemia threshold \(T_{hyper}\) indicate excessively high levels. Measurements that fall between these two thresholds are considered within the normal range. In practice, \(T_{hypo}\) and \(T_{hyper}\) are often set to 70 mg/dL and 180 mg/dL, respectively, based on guidelines and prior studies \cite{chico2024hybrid, marigliano2024glucose, vettoretti2020advanced}.

Prediction errors in the critical regions of hypo- and hyperglycemia are particularly dangerous for patients with T1D, as they can lead to missed interventions and serious complications. Woldaregay et al.~\cite{woldaregay2019data} highlighted that prior studies rarely account for the uneven clinical impact of such errors. To address this gap, we propose penalizing prediction errors more heavily in critical regions than in the normal range.

\begin{equation}
\text{Glucose Level Regions} = 
\begin{cases}
\text{hypoglycemia}, & \quad \phantom{T_{hypo} \leq{}} x < T_{hypo} \\
\text{normal}, & \quad T_{hypo} \leq x \leq T_{hyper} \\
\text{hyperglycemia}, & \quad \phantom{T_{hypo} \leq{}} x > T_{hyper}
\label{eq:glucose_region_class}
\end{cases}
\end{equation}

We then break down the total error, which represents the cumulative prediction error across all time steps, into separate components computed within each of the three glucose level regions defined in (\ref{eq:glucose_region_class}). Accordingly, the total error is expressed as the aggregation of regional errors corresponding to hypoglycemia, normal, and hyperglycemia. By assigning region-specific weights, the formulation places greater emphasis on prediction errors occurring in the critical hypo- and hyperglycemic ranges while maintaining lower sensitivity within the normal region. The total error is computed as the weighted summation of these regional errors, as formalized in (\ref{eq}).

\begin{equation} 
\label{eq}
{Error}_{total} = \sum_{i=1}^{3} w_i \times {Error}_{i} 
\end{equation}

The total error serves as the model’s loss function aggregated over all instances, enabling it to capture and prioritize prediction accuracy within distinct regions. If we choose Mean Absolute Error (MAE) as the error term \({Error}_i\), and assign weights \(w_{hypo}\), \(w_{normal}\), and \(w_{hyper}\) to the hypoglycemia, normal, and hyperglycemia regions, respectively, the total error can be calculated as follows:

\begin{equation}
{MAE} = \frac{1}{N} \sum_{i=1}^{N} \left| y_i - \hat{y_i} \right|
\label{eq:mae_formula}
\end{equation}
\begin{equation}
\begin{aligned}
{Error}_{total} &=
\frac{w_{hypo}}{n_1} \phantom{i} \sum_{i=1}^{n_1} \left|y_i - \hat{y}_i \right| + 
\frac{w_{normal}}{n_2} \sum_{i=1}^{n_2} \left|y_i - \hat{y}_i \right| +
\frac{w_{hyper}}{n_3} \sum_{i=1}^{n_3} \left|y_i - \hat{y}_i \right|
\label{eq:custom_loss_function}
\end{aligned}
\end{equation}

\noindent where \(n_1\), \(n_2\) and \(n_3\) represent the total number of blood glucose samples in each region, and \(y_i\) and \(\hat{y}_i\) represent the predicted value and true value of the glucose level, respectively. The main research question that remains to be answered is how to determine the parameters of this error equation, including weight and threshold values.

\subsection{Error Weights}

We classify CGM values into regions based on the thresholds discussed earlier. Next, we finalize the custom loss function by determining the optimal weights for each region, with emphasis on dysglycemia. To simplify the optimization, we fix $w_{normal} = 1$, reducing the search to two continuous parameters, $w_{hypo}$ and $w_{hyper}$. Estimating these weights constitutes a general optimization problem in which the objective is defined by the prediction error obtained after model training. 

In this study, we employ a Genetic Algorithm (GA) to search for suitable weight values because it provides a simple, gradient-free procedure for continuous parameter spaces where each candidate solution must be evaluated through model training and validation \cite{immanuel2019genetic, jin2005comprehensive, khamesian2024hybrid}. Within the GA, the fitness of each candidate weight pair is defined as the validation RMSE of the forecasting model, such that lower prediction error corresponds to higher fitness. Although alternative optimization strategies such as grid search or random search can also be applied to this problem, we selected the GA due to its flexibility and efficiency in continuous search spaces. A comparative evaluation of GA, grid search, and random search is presented in Section~\ref{sec:weights_optimization}, where the GA consistently achieved lower validation error. The resulting procedure optimizes $w_{hypo}$ and $w_{hyper}$ separately for each patient, aligning the loss function with the clinical importance of different glucose regions. The optimization steps are summarized in Algorithm~\ref{alg:genetic_algorithm}.

\begin{algorithm}[!t]
\renewcommand{\baselinestretch}{1.2}
\caption{Finding the Best Pair of Weights for Each Patient}
\label{alg:genetic_algorithm}
\small
\begin{algorithmic}[1]
\STATE \textbf{Parameters:}
\STATE $G \gets 25$ \tcp{Number of Generations}
\STATE $N \gets 20$ \tcp{Population Size}

\FOR{each patient}
    \STATE Initialize population $P$ with random weight vectors $w$
    \STATE $w = [w_{hypo}, w_{hyper}]^T \in \mathbb{R}^2$
    \STATE $w_{hypo}, w_{hyper} \sim \mathcal{U}(1, 10)$
    
    \FOR{$g \gets 1$ \TO $G$}
        \FOR{each individual $i \in P$}
            \STATE evaluate fitness $f_i$
        \ENDFOR
        
        \STATE sort population $P$ by fitness
        \STATE $P_{\text{best}} \gets \text{top } N/2 \text{ individuals from } P$
        
        \STATE $P_{\text{offspring}} \gets [\,]$
        
        \WHILE{$|P_{\text{offspring}}| < N/2$}
            \STATE randomly select parents $p_1, p_2 \in P_{\text{best}}$
            \STATE $c \gets \frac{1}{2}(p_1 + p_2)$
            \STATE $m \sim \mathcal{N}(0, 0.5)$
            \STATE $c \gets \text{clip}(c + m, 1, 10)$
            \STATE $P_{\text{offspring}} \gets P_{\text{offspring}} \cup c$
        \ENDWHILE
        
        \STATE $P \gets P_{\text{best}} \cup P_{\text{offspring}}$
    \ENDFOR
    
    \STATE $w^* \gets \arg\min(f_i)$
    \STATE save $w^*$
\ENDFOR

\end{algorithmic}
\end{algorithm}

\begin{figure}[!t]
\centerline{\includegraphics[width=\textwidth]{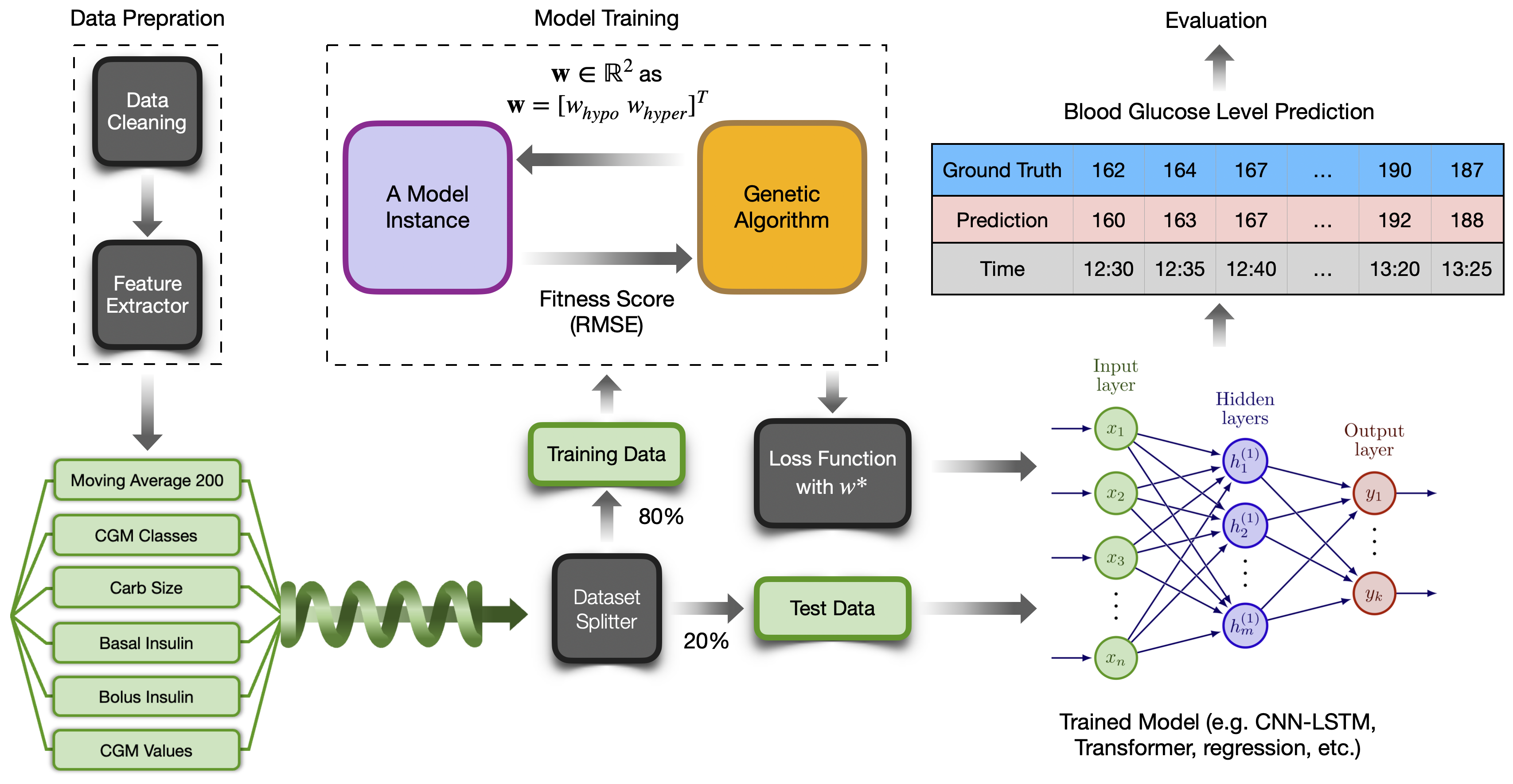}}
\captionsetup{font=small}
\caption{The proposed GLIMMER methodology for predicting blood glucose levels in patients with T1D. CGM data undergo preprocessing and feature extraction before being split for training and testing. A genetic algorithm optimizes a custom loss function used to train an architecture-agnostic prediction model (e.g., CNN-LSTM, Transformer), which is then evaluated on the test data using the finalized parameters.}
\label{fig:2}
\end{figure}

In summary, our methodology combines three key components to enhance prediction performance: selecting effective input features, designing a custom loss function that emphasizes clinically critical regions, and using a genetic algorithm to tune the loss function’s weights. Fig.~\ref{fig:2} illustrates the complete methodology for the design and evaluation of GLIMMER.

\section{Data}\label{sec:data}
\subsection{OhioT1DM Dataset}
We utilized the OhioT1DM dataset \cite{marling2020ohiot1dm}, which includes data from 12 individuals with T1D. The dataset contains raw glucose values recorded every 5 minutes, along with basal insulin, bolus insulin, and carbohydrate intake over an 8-week period.

\subsection{AZT1D Dataset}
Alongside the OhioT1DM dataset, we introduced and published a new dataset named AZT1D~\cite{khamesian2025azt1d}. We collected data from 25 patients with T1D using AID systems who visited the endocrinology clinic at the Mayo Clinic in Scottsdale, AZ, between December 2023 and April 2024 for their regular appointments. The participants included 13 females and 12 males, aged between 27 and 80 years, with an average age of 59 years. Informed consent was obtained from all participants under the study protocol (IRB \#23-003065). For each patient, the dataset contains an average of one month of real-world recordings, including CGM signals captured with Dexcom G6 Pro, insulin logs, meal carbohydrate sizes, and device modes (regular/sleep/exercise) recorded from the Tandem t:slim X2 insulin pump. In total, the dataset comprises 320{,}488 CGM entries spanning approximately 26{,}707 hours of monitoring data.

\subsection{Data Preparation}
The OhioT1DM dataset consists of 24 files from 12 patients, with two files per patient: one for training and one for testing \cite{marling2020ohiot1dm}. To avoid overfitting and to enable hyperparameter tuning, we further split the training file chronologically, using the first 80\% for training and the remaining 20\% for validation. The separate testing file is kept intact and used exclusively for final evaluation. For the AZT1D dataset \cite{khamesian2025azt1d}, which does not come pre‑partitioned, we applied a similar strategy: each patient’s data was split into 80\% for training and 20\% for testing, and then the training portion was again split into 80\% training and 20\% validation.

In addition to CGM values, basal insulin, bolus insulin, and carbohydrate amounts, we crafted two additional features. First, we computed a 200-point moving average of CGM values to capture longer-term glucose trends while reducing sensitivity to short-term fluctuations caused by meals, physical activity, or stress. Moving averages are commonly used in time-series forecasting to smooth noise and emphasize underlying trends, thereby improving predictive stability and generalization \cite{zheng2018feature, makridakis1994time}. With five-minute sampling, 200 points correspond to approximately 16–17 hours. We adopted this long-horizon window—analogous to the widely used 200-period moving average for trend extraction in noisy time series—to provide extended temporal context while keeping the input representation simple and without increasing architectural complexity. Second, we labeled each CGM value as hypoglycemic, normal, or hyperglycemic based on clinical thresholds (see Section~\ref{sec:metholodogy}) to help the model distinguish clinically critical regions. Overall, all six features were engineered during the data preprocessing stage and prepared as inputs to the model.

\section{Experimental Setup}
\label{sec:experimental_setup}

\subsection{Architecture Configuration}
To demonstrate that GLIMMER supports an architecture-agnostic design, we implemented and compared two model types: a CNN-LSTM architecture and a Transformer-based architecture. The CNN-LSTM model includes three 1D convolutional layers for local feature extraction, followed by a single LSTM layer with 8 units, reflecting configurations commonly used in prior studies~\cite{el2020deep, dylag2023machine}. As an alternative, we implemented a CNN-Transformer architecture which uses the same convolutional front-end, followed by a single Transformer block with 8 attention heads, a key dimension of 16, a feed-forward size of 256, and a dropout rate of 0.1. This design enables a fair comparison of different architectural backbones under a unified training and evaluation pipeline.

\subsection{Custom Loss Function Parameters Configuration}
\label{sec:weights_optimization}

We first compared GA with grid search and random search for selecting the custom loss weights. For each forecasting model, weight selection was evaluated on the OhioT1DM validation set using RMSE as the objective metric. As reported in Table~\ref{tab:2}, GA achieved lower mean validation RMSE than both grid search and random search for the CNN-LSTM and Transformer models, indicating improved effectiveness in the continuous weight space.

\begin{table}[b]
\centering
\captionsetup{font=small}
\small
\renewcommand{\arraystretch}{1.2}
\caption{Comparison of grid search, random search, and GA for selecting the custom loss weights on the OhioT1DM validation set. For each model, values represent Mean $\pm$ STD of RMSE (mg/dL) computed across patients.}

\begin{tabular}{lccc}
\toprule
\textbf{Model} & \textbf{Grid Search} & \textbf{Random Search} & \textbf{Genetic Algorithm} \\
\midrule
GLIMMER (CNN-LSTM)     & $27.51 \pm 3.94$ & $26.70 \pm 3.72$ & \textbf{24.47 $\pm$ 4.05} \\
GLIMMER (Transformer)  & $31.39 \pm 4.55$ & $32.91 \pm 7.43$ & \textbf{29.94 $\pm$ 4.31} \\
\bottomrule
\end{tabular}
\label{tab:2}
\end{table}

Following this comparative analysis, GA was applied independently to each patient in the OhioT1DM training set. For every patient, the pair $(w_{\text{hypo}}, w_{\text{hyper}})$ was optimized by minimizing validation RMSE. The resulting patient-specific weights are summarized in Table~\ref{tab:3} together with their Mean $\pm$ STD across patients. For evaluation on the test set, the weights were fixed to the average values obtained across patients. Specifically, the CNN-LSTM model used $(3.29, 2.38)$, whereas the Transformer model used $(4.67, 1.71)$. These averaged weights provide a stable configuration while preserving the dysglycemia emphasis learned during the patient-specific optimization phase.

\begin{table}[b]
\centering
\captionsetup{font=small}
\small
\renewcommand{\arraystretch}{1.2}
\caption{Patient-specific dysglycemia weights ($w_{\text{hypo}}$, $w_{\text{hyper}}$) obtained using Genetic Algorithm optimization on the OhioT1DM validation set. Weights were selected by minimizing validation RMSE for each forecasting model. Values are reported per patient and summarized as Mean $\pm$ Population STD.}

\begin{tabular}{c cc cc}
\toprule
\multirow{2}{*}{Patient ID} 
& \multicolumn{2}{c}{\textbf{GLIMMER (CNN-LSTM)}} 
& \multicolumn{2}{c}{\textbf{GLIMMER (Transformer)}} \\
\cmidrule(lr){2-3} \cmidrule(lr){4-5}
& $w_{\text{hypo}}$ & $w_{\text{hyper}}$ 
& $w_{\text{hypo}}$ & $w_{\text{hyper}}$ \\
\midrule
559 & 2.63 & 1.44 & 3.16 & 1.10 \\
563 & 2.55 & 1.00 & 3.40 & 1.10 \\
570 & 2.46 & 1.00 & 4.91 & 1.55 \\
575 & 5.18 & 2.67 & 6.09 & 1.87 \\
588 & 1.17 & 3.52 & 6.01 & 2.63 \\
591 & 1.00 & 2.89 & 3.72 & 1.90 \\
540 & 4.17 & 2.53 & 5.09 & 1.53 \\
544 & 5.69 & 2.76 & 6.57 & 2.65 \\
552 & 3.66 & 3.09 & 4.52 & 2.53 \\
567 & 5.58 & 1.32 & 4.37 & 1.28 \\
584 & 1.64 & 1.60 & 4.21 & 1.30 \\
596 & 3.78 & 4.72 & 4.05 & 1.17 \\
\midrule
\textbf{Mean $\pm$ STD} & \textbf{3.29 $\pm$ 1.61} & \textbf{2.38 $\pm$ 1.05} & \textbf{4.67 $\pm$ 1.09} & \textbf{1.71 $\pm$ 0.58} \\
\bottomrule
\end{tabular}
\label{tab:3}
\end{table}

\subsection{Evaluation Metrics}
To assess the performance of the GLIMMER framework, we employed standard metrics commonly used in related studies \cite{zhu2022personalized, li2019convolutional, aliberti2019multi, butt2023feature}, including:

\subsubsection{Root Mean Square Error (RMSE)}
RMSE quantifies the square root of the average squared differences between predicted and actual values:
\begin{equation}
\text{RMSE} = \sqrt{\frac{1}{n} \sum_{i=1}^{n} (y_i - \hat{y}_i)^2}
\end{equation}
where \( y_i \) and \( \hat{y}_i \) are the actual and predicted CGM values, respectively.

\subsubsection{Mean Absolute Error (MAE)}
MAE measures the average magnitude of prediction errors:
\begin{equation}
\text{MAE} = \frac{1}{n} \sum_{i=1}^{n} |y_i - \hat{y}_i|
\end{equation}

\subsubsection{Precision, Recall, and F1 Score}
These metrics evaluate event-level detection (e.g., hypoglycemia onset). 
Precision measures the proportion of correctly predicted positive events among all predicted positives, whereas recall measures the proportion of correctly predicted positive events among all actual positives. The F1 score provides a harmonic mean of precision and recall.

Let TP denote true positives (correctly detected events), FP denote false positives (incorrectly predicted events), and FN denote false negatives (missed events). The metrics are defined as:

\begin{equation}
\text{Precision} = \frac{TP}{TP + FP}, \quad
\text{Recall} = \frac{TP}{TP + FN}, \quad
\text{F1} = 2 \cdot \frac{\text{Precision} \cdot \text{Recall}}{\text{Precision} + \text{Recall}}
\end{equation}

\subsubsection{Clarke Error Grid Analysis}
Clarke Error Grid (CEG)~\cite{clarke1987evaluating} evaluates clinical relevance by categorizing prediction-reference pairs into five zones:
\begin{itemize}
    \item \textbf{A}: Accurate predictions (within 20\% of reference)
    \item \textbf{B}: Benign errors (no risk of inappropriate treatment)
    \item \textbf{C}: Unnecessary treatment
    \item \textbf{D}: Dangerous missed hypo-/hyperglycemia
    \item \textbf{E}: Incorrect treatment (confusing hypo/hyper)
\end{itemize}

\section{Results}
\subsection{GLIMMER Performance on Baseline Architectures}
Table~\ref{tab:4} compares the performance of CNN-LSTM and Transformer architectures before and after applying the GLIMMER framework on the OhioT1DM and AZT1D datasets, each with a 60-minute prediction horizon (PH). Across both datasets and models, GLIMMER consistently improves performance in glucose forecasting and dysglycemia detection. On average, GLIMMER reduces RMSE by 24.6\% and MAE by 29.6\% for CNN-LSTM, and by 8.4\% and 11.6\% for Transformer. For dysglycemia classification, GLIMMER boosts recall to 98.4\% and F1-score to 86.8\% in the CNN-LSTM model, with similar improvements observed in the Transformer variant. These results highlight GLIMMER’s ability to enhance both predictive accuracy and clinical reliability across diverse models and datasets.

\begin{table}[t]
\centering
\captionsetup{font=small}
\scriptsize
\setlength{\tabcolsep}{3.5pt}
\renewcommand{\arraystretch}{1.2}
\caption{Performance comparison of CNN-LSTM and Transformer architectures before and after applying the GLIMMER framework across two datasets (PH = 60 min). Metrics include RMSE and MAE for glucose prediction, and Recall, Precision, and F1 for dysglycemia.}

\begin{tabular}{lcccccc}
\toprule
Model & RMSE & MAE & Recall & Precision & F1 \\ & (mg/dL) & (mg/dL) & (\%) & (\%) & (\%) \\
\midrule
\multicolumn{6}{c}{\textbf{OhioT1DM Dataset with PH = 60 min}} \\
\midrule
CNN-LSTM Baseline      & 31.98 $\pm$ 4.15 & 23.00 $\pm$ 2.87 & 87.87 & 74.56 & 80.67 \\
GLIMMER(CNN-LSTM)      & 23.97 $\pm$ 3.77 & 15.83 $\pm$ 2.09 & 99.78 & 84.66 & 91.60 \\
\midrule
Transformer Baseline   & 30.08 $\pm$ 3.34 & 22.48 $\pm$ 2.47 & 89.09 & 70.77 & 78.88 \\
GLIMMER(Transformer)   & 27.96 $\pm$ 4.02 & 20.17 $\pm$ 2.77 & 97.86 & 72.13 & 83.04 \\
\midrule
\multicolumn{6}{c}{\textbf{AZT1D Dataset with PH = 60 min}} \\
\midrule
CNN-LSTM Baseline      & 29.55 $\pm$ 6.49 & 21.61 $\pm$ 5.19 & 78.33 & 45.24 & 57.35 \\
GLIMMER(CNN-LSTM)      & 22.48 $\pm$ 3.57 & 15.58 $\pm$ 2.87 & 96.97 & 71.14 & 82.07 \\
\midrule
Transformer Baseline   & 28.15 $\pm$ 4.94 & 21.27 $\pm$ 3.99 & 79.87 & 66.11 & 72.35 \\
GLIMMER(Transformer)   & 25.39 $\pm$ 5.21 & 18.50 $\pm$ 4.03 & 89.66 & 63.17 & 74.12 \\
\midrule
\multicolumn{6}{c}{\textbf{Average}} \\
\midrule
CNN-LSTM Baseline      & 30.77 $\pm$ 5.32 & 22.31 $\pm$ 4.03 & 83.10 & 59.90 & 69.01 \\
GLIMMER(CNN-LSTM)      & \textbf{23.22 $\pm$ 3.67} & \textbf{15.71 $\pm$ 2.48} & \textbf{98.38} & \textbf{77.90} & \textbf{86.84} \\
Transformer Baseline   & 29.12 $\pm$ 4.14 & 21.88 $\pm$ 5.22 & 84.48 & 68.44 & 75.62 \\
GLIMMER(Transformer)   & 26.68 $\pm$ 4.61 & 19.34 $\pm$ 3.40 & 93.76 & 67.65 & 78.58 \\
\bottomrule
\end{tabular}
\label{tab:4}
\end{table}

While the results indicate consistent improvements across architectures and datasets, statistical validation is necessary to confirm that these gains are not attributable to random variation. We conducted a paired Wilcoxon signed-rank test ~\cite{rosner2006wilcoxon} using the per-patient RMSE values for each baseline model and its corresponding GLIMMER-enhanced version. The Wilcoxon test was selected because it is a non-parametric paired test that does not assume normality and is appropriate for comparing matched samples (i.e., the same patients evaluated before and after applying GLIMMER). For each dataset and architecture, we formed patient-wise RMSE pairs and tested the null hypothesis that the median difference between baseline and GLIMMER models is zero. As reported in Table~\ref{tab:wilcoxon}, all comparisons achieved statistical significance at $\alpha = 0.05$. On the OhioT1DM dataset, p-values were $4.88 \times 10^{-4}$ for CNN-LSTM and $4.24 \times 10^{-2}$ for Transformer. On the AZT1D dataset, the improvements were even more pronounced, with p-values of $5.96 \times 10^{-8}$ and $5.32 \times 10^{-5}$, respectively. These findings confirm that the improvements introduced by GLIMMER are consistent across patients and unlikely to be explained by random variation.

\begin{table}[t]
\centering
\captionsetup{font=small}
\small
\setlength{\tabcolsep}{20pt}
\renewcommand{\arraystretch}{1.2}
\caption{Results of the paired Wilcoxon signed-rank test comparing baseline models and their GLIMMER-enhanced versions using per-patient RMSE values across each dataset.}
\label{tab:wilcoxon}
\begin{tabular}{lcc}
\toprule
Model & p-value & Significant ($\alpha=0.05$) \\
\midrule
\multicolumn{3}{c}{\textbf{OhioT1DM Dataset}} \\
\midrule
CNN-LSTM & $4.88 \times 10^{-4}$ & Yes \\
Transformer & $4.24 \times 10^{-2}$ & Yes \\
\midrule
\multicolumn{3}{c}{\textbf{AZT1D Dataset}} \\
\midrule
CNN-LSTM & $5.96 \times 10^{-8}$  & Yes \\
Transformer & $5.32 \times 10^{-5}$ & Yes \\
\bottomrule
\end{tabular}
\end{table}

\begin{figure}[!t]
\centering
\captionsetup{font=small}

\begin{subfigure}[t]{0.87\textwidth}
    \centering
    \includegraphics[width=\textwidth]{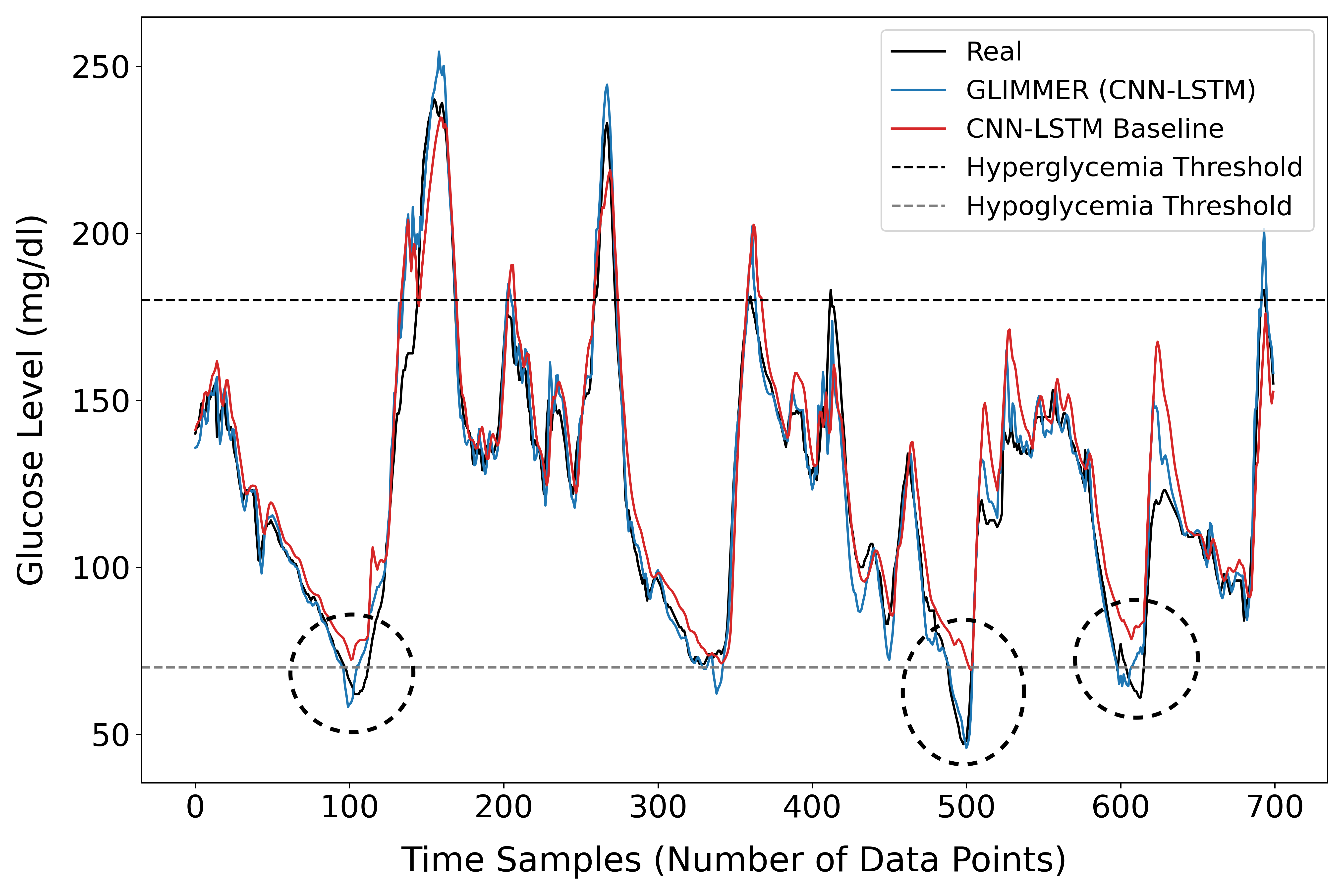}
    \caption{CNN-LSTM vs. GLIMMER (CNN-LSTM)}
\end{subfigure}

\begin{subfigure}[t]{0.87\textwidth}
    \centering
    \includegraphics[width=\textwidth]{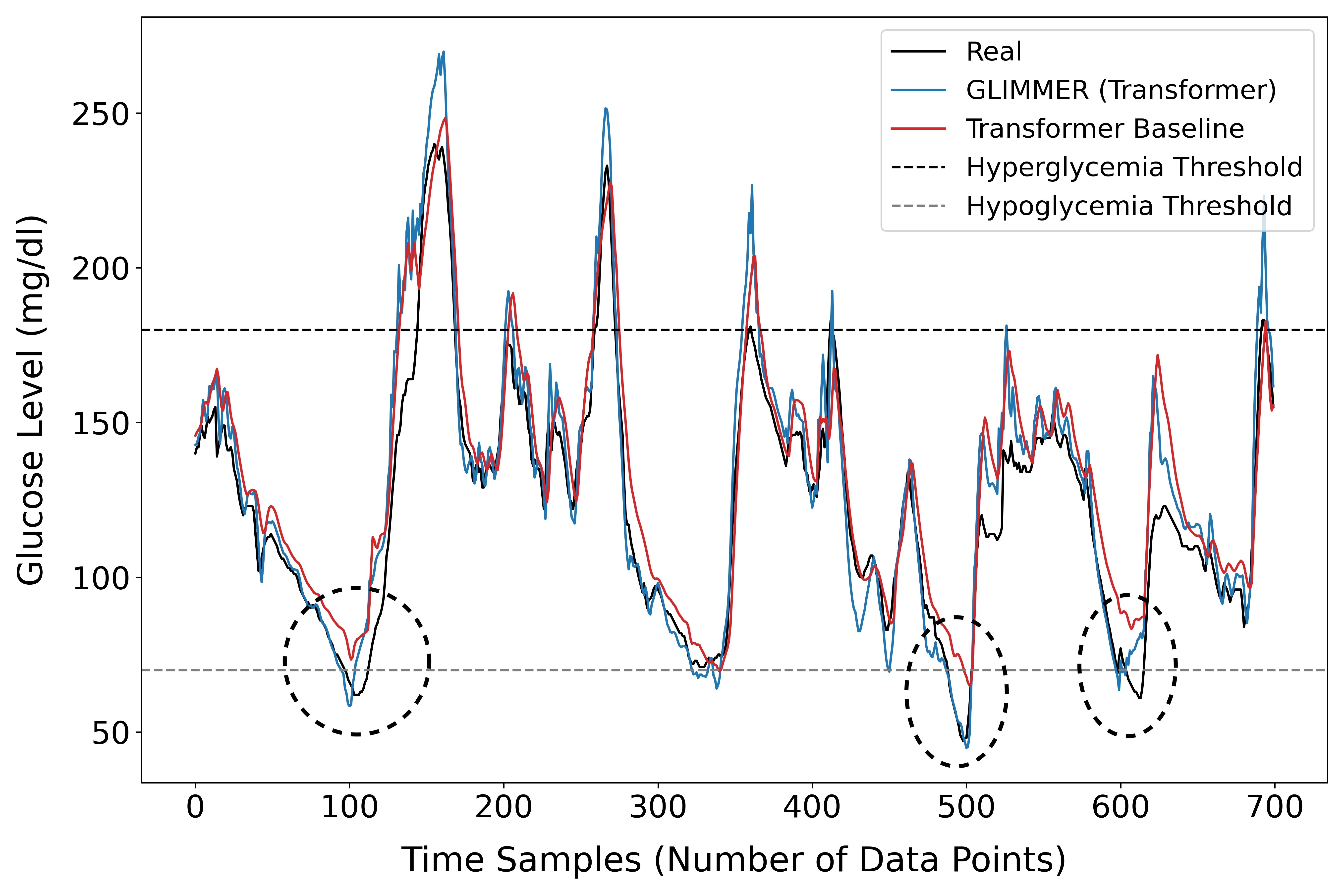}
    \caption{Transformer vs. GLIMMER (Transformer)}
\end{subfigure}

\caption{Forecasting comparison between baseline models and their GLIMMER-enhanced versions for patient 552 from the OhioT1DM test dataset. The solid black line represents the real CGM values. Dashed black and gray lines mark the hyperglycemia and hypoglycemia thresholds. The black circles highlight dysglycemic events that were correctly detected by GLIMMER but missed by the respective baseline models.}
\label{fig:3}
\end{figure}

To better illustrate the improved performance reported in Table~\ref{tab:4}, Fig.~\ref{fig:3} presents a case study of patient 552 from the OhioT1DM dataset. Subfigure (a) shows the comparison between the baseline CNN-LSTM and its GLIMMER-enhanced version, while subfigure (b) presents the same comparison for the Transformer architecture. In both cases, GLIMMER successfully detects three hypoglycemia events, whereas the respective baseline models fail to capture these critical excursions. These regions, marked by red ellipses, highlight GLIMMER’s improved ability to forecast dysglycemic events and are consistent with the higher recall and F1 scores observed in safety-critical zones.

\subsection{Clinical Validity of GLIMMER-Enhanced Models}
\begin{figure}[htbp]
\centerline{\includegraphics[angle=90, width=0.75\textwidth]{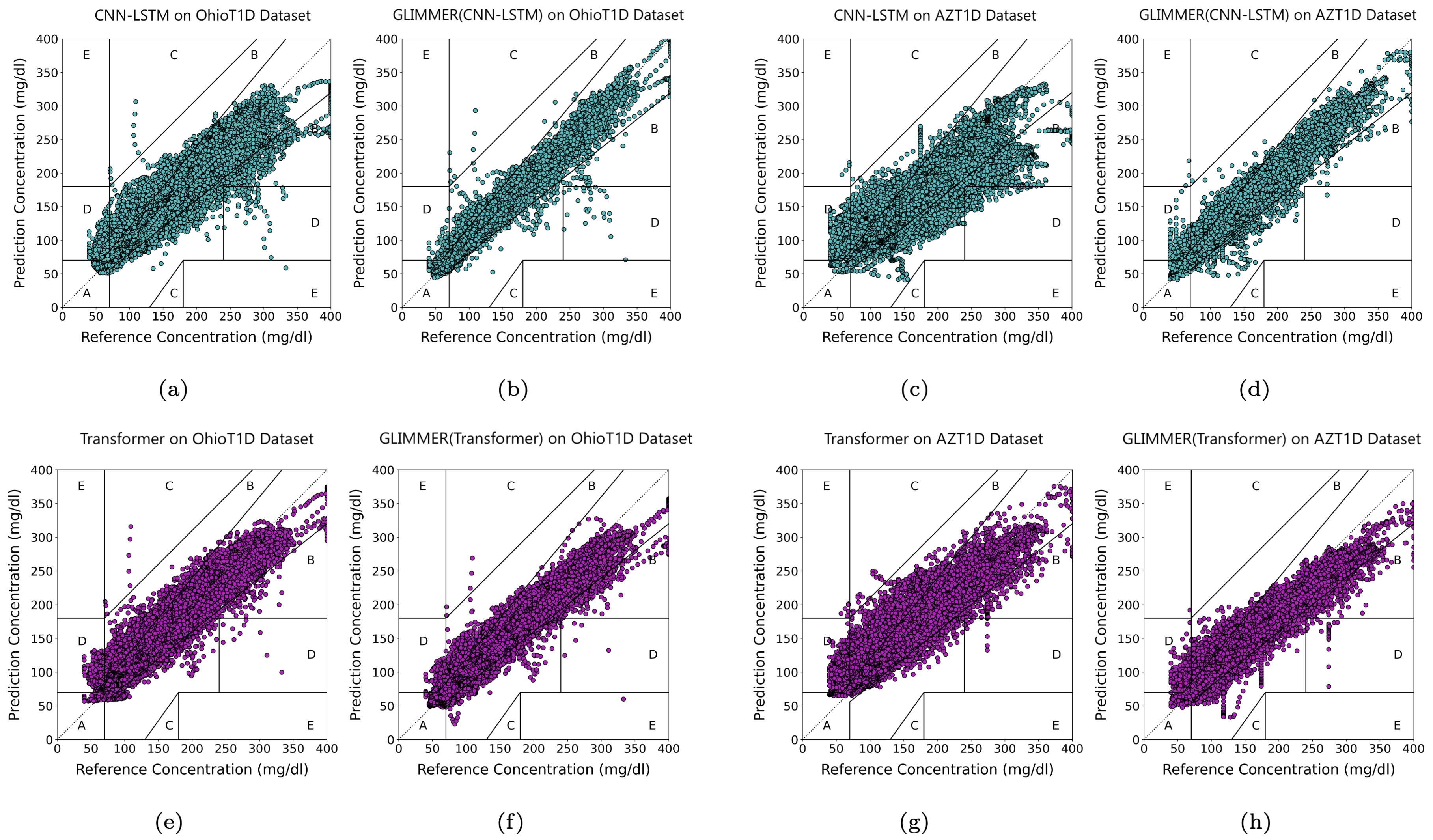}}
\captionsetup{font=small}
\caption{Clarke Error Grid for CNN-LSTM and Transformer models, with and without GLIMMER, across both datasets. Subfigures (a)–(d) correspond to the OhioT1DM and AZT1D datasets using the CNN-LSTM baseline model and its GLIMMER-enhanced version, respectively. Subfigures (e)–(h) show the same comparison using the Transformer architecture. In each case, GLIMMER improves the clustering of predictions near the x=y line.}
\label{fig:4}
\end{figure}

To further assess the clinical reliability of glucose predictions, we conducted a CEG analysis, which classifies prediction errors into zones based on their potential clinical consequences. As shown in Table~\ref{tab:5} and visualized in Fig.~\ref{fig:4}, GLIMMER consistently increases the proportion of predictions falling within Zone~A—representing clinically accurate and safe predictions—while reducing the presence of errors in critical-risk zones (D and E). In the OhioT1DM dataset, GLIMMER improves Zone~A coverage from 74.3\% to 85.5\% for CNN-LSTM, and from 72.8\% to 79.7\% for Transformer, with corresponding reductions in Zones~D and~E. In AZT1D, Zone~A rises from 73.3\% to 83.9\% for CNN-LSTM and from 73.0\% to 80.3\% for Transformer, with Zone~D errors nearly halved or more. Averaged across datasets, CNN-LSTM with GLIMMER achieves a 10.9 percentage point gain in Zone~A and a reduction of over 50\% in Zone~D errors. Transformer shows a 7-point improvement in Zone~A and similarly low rates of high-risk misclassifications. These results demonstrate that GLIMMER not only enhances dysglycemia detection and predictive accuracy, but also significantly improves clinical safety by reducing the likelihood of dangerous prediction errors.

\begin{table}[t]
\centering
\captionsetup{font=small}
\scriptsize
\setlength{\tabcolsep}{10.15pt}
{\renewcommand{\arraystretch}{1.2}
\caption{CEG zone distribution (\%) for glucose predictions made by CNN-LSTM and Transformer architectures, before and after applying GLIMMER, on the OhioT1DM and AZT1D datasets (PH = 60 minutes). Zone~A reflects clinically accurate predictions, while Zones~D and~E indicate potentially dangerous errors.}
\begin{tabular}{lccccc}
\toprule
Model & Zone A & Zone B & Zone C & Zone D & Zone E \\
\midrule
\multicolumn{6}{c}{\textbf{OhioT1DM Dataset with PH = 60 min}} \\
\midrule
CNN-LSTM Baseline      & 74.31 & 23.12 & 0.11 & 2.43 & 0.03 \\
GLIMMER(CNN-LSTM)      & 85.46 & 13.26 & 0.13 & 1.12 & 0.02 \\
\midrule
Transformer Baseline   & 72.78 & 24.08 & 0.14 & 2.99 & 0.00 \\
GLIMMER(Transformer)   & 79.67 & 18.50 & 0.07 & 1.72 & 0.04 \\
\midrule
\multicolumn{6}{c}{\textbf{AZT1D Dataset with PH = 60 min}} \\
\midrule
CNN-LSTM Baseline      & 73.27 & 24.47 & 0.03 & 2.21 & 0.02 \\
GLIMMER(CNN-LSTM)      & 83.89 & 14.94 & 0.02 & 1.12 & 0.02 \\
\midrule
Transformer Baseline   & 73.01 & 24.92 & 0.04 & 2.00 & 0.02 \\
GLIMMER(Transformer)   & 80.26 & 18.17 & 0.01 & 1.54 & 0.02 \\
\midrule
\multicolumn{6}{c}{\textbf{Average}} \\
\midrule
CNN-LSTM Baseline       & 73.79 & 23.80 & 0.07 & 2.32 & 0.03 \\
GLIMMER(CNN-LSTM)       & \textbf{84.68} & \textbf{14.10} & 0.08 & \textbf{1.12} & 0.02 \\
Transformer Baseline    & 72.90 & 24.50 & 0.09 & 2.50 & \textbf{0.01} \\
GLIMMER(Transformer)    & 79.97 & 18.34 & \textbf{0.04} & 1.63 & 0.03 \\
\bottomrule
\label{tab:5}
\end{tabular}}
\end{table}

\subsection{Benchmarking Model Performance and Complexity}
Table~\ref{tab:6} benchmarks various glucose forecasting models on the OhioT1DM dataset using RMSE, parameter count, and whether a weighted loss formulation was employed. While top-performing models such as TimesNet~\cite{wu2022timesnet}, BG-BERT~\cite{zheng2024predicting}, and Gluformer~\cite{sergazinov2023gluformer} achieve strong predictive accuracy, they require a substantial number of parameters—ranging from 2 million to nearly 19 million—and typically depend on powerful GPU hardware for training and inference. In contrast, GLIMMER(CNN-LSTM) achieves a competitive RMSE of 23.97~mg/dL while using only 10,000 parameters—over \textbf{1,800$\times$ fewer} than TimesNet and \textbf{200$\times$ fewer} than BG-BERT.In addition to architectural comparisons, it is important to consider methods that incorporate weighted or penalty-based loss formulations. Among these approaches—such as Cichosz et al.~\cite{cichosz2021penalty}, SVR-XGBoost~\cite{katsarou2025optimizing}, FedGlue~\cite{dave2024fedglu}, BG-BERT~\cite{zheng2024predicting}, and Sequential-T~\cite{barbato2025lightweight}—the lowest reported RMSE is 23.67~mg/dL (BG-BERT). GLIMMER(CNN-LSTM) achieves a closely comparable RMSE of 23.97~mg/dL, outperforming other weighted-loss approaches such as FedGlue (29.25~mg/dL), Sequential-T (28.99~mg/dL), Cichosz et al. (34.33~mg/dL), and SVR-XGBoost (35.94~mg/dL). These results indicate that the proposed loss design enables competitive accuracy among weighted-loss strategies without relying on increased architectural complexity. Overall, this finding highlights GLIMMER’s ability to approach the performance of large transformer-based models with drastically lower computational cost, making it far more suitable for real-time deployment in mobile or resource-constrained clinical environments. By optimizing baseline architectures through the GLIMMER framework, we demonstrate that it is possible to achieve both high predictive performance and lightweight design—an essential combination for integration into diabetes management tools such as AID systems.

\begin{table}[t]
\centering
\captionsetup{font=small}
\scriptsize
\renewcommand{\arraystretch}{1.5}
\caption{Comparison of RMSE and parameter count of glucose forecasting models on the OhioT1DM dataset (PH = 60 minutes).}
\setlength{\tabcolsep}{8pt}
\begin{tabular}{lccc}
\toprule
\textbf{Model} & \textbf{RMSE (mg/dL)} & \textbf{Params} & \textbf{Weighted Loss?}\\
\midrule
FCNN (2022)~\cite{zhu2022personalized}                 & 31.07 $\pm$ 3.62 & N/A       & No  \\
CRNN (2019)~\cite{li2019convolutional}                 & 32.02 $\pm$ 3.76 & N/A       & No  \\
Bi-LSTM (2018)~\cite{sun2018predicting}                & 33.44 $\pm$ 3.76 & N/A       & No  \\
Informer (2021)~\cite{zhou2021informer}                & 25.76 $\pm$ N/A  & 181k      & No  \\
TimesNet (2022)~\cite{wu2022timesnet}                  & 23.48 $\pm$ N/A  & 18,749k   & No  \\
Gluformer (2023)~\cite{sergazinov2023gluformer}        & 27.14 $\pm$ N/A  & 11,247k   & No  \\
Cichosz et al. (2021)~\cite{cichosz2021penalty}        & 34.33 $\pm$ N/A  & N/A       & Yes \\
SVR-XGBoost (2025)~\cite{katsarou2025optimizing}       & 35.94 $\pm$ 5.19 & N/A       & Yes \\
FedGlue (2024)~\cite{dave2024fedglu}                   & 29.25 $\pm$ N/A  & 160k      & Yes \\
BG-BERT (2024)~\cite{zheng2024predicting}              & 23.67 $\pm$ N/A  & 2,091k    & Yes \\
Sequential-T (2025)~\cite{barbato2025lightweight}      & 28.99 $\pm$ N/A  & 123k      & Yes \\
\midrule
CNN-LSTM Baseline                               & 31.98 $\pm$ 4.15 & 10k      & No  \\
GLIMMER(CNN-LSTM)                               & 23.97 $\pm$ 3.77 & 10k      & Yes  \\
Transformer Baseline                            & 30.08 $\pm$ 3.34 & 50k      & No  \\
GLIMMER(Transformer)                            & 27.96 $\pm$ 4.02 & 50k      & Yes  \\
\bottomrule
\end{tabular}
\label{tab:6}
\end{table}

\section{Discussion}
This study introduces GLIMMER as a modular, architecture-agnostic training framework that integrates structured feature engineering, region-aware loss formulation, and data-driven weight optimization to improve glucose forecasting in type 1 diabetes. Unlike conventional forecasting pipelines that rely on uniform error objectives, GLIMMER explicitly models the unequal clinical impact of prediction errors across glycemic regions and formulates regional weight selection as a continuous optimization problem. By learning dysglycemia-sensitive weights using a genetic algorithm rather than relying on fixed or heuristic penalties, the framework replaces manual tuning with systematic calibration while remaining independent of the underlying architecture. Applied to both CNN-LSTM and Transformer models, GLIMMER consistently reduced RMSE and MAE, improved dysglycemia detection, and enhanced Clarke Error Grid distributions across two independent datasets, including the newly introduced AZT1D cohort. The consistent improvements observed across fundamentally different backbone architectures demonstrate that the framework operates independently of model structure rather than being tailored to a specific network design. The paired Wilcoxon analysis further confirmed that these improvements were statistically significant across patients.

The observed improvements in recall and F1-score for dysglycemia detection are clinically meaningful. Higher recall indicates that more true hypo- and hyperglycemic events are correctly anticipated, reducing the risk of missed adverse events. Although increased recall can reduce precision and lead to additional false alarms, the modest decline in precision suggests improved sensitivity without substantial loss of specificity. In safety-critical applications, under-detection of adverse events is generally more harmful than moderate over-detection. This clinical benefit is further supported by Clarke Error Grid analysis, which shows increased Zone A predictions and reduced Zone D errors, demonstrating that emphasizing dysglycemic regions during training enhances not only statistical performance but also clinical safety.

Despite these promising findings, several limitations should be acknowledged. First, hypoglycemic events remain relatively sparse in both datasets, which limits the model’s exposure to low-glucose patterns during training. Although weighted loss partially compensates for this imbalance, performance in rare-event regions may still be constrained by data availability. Second, weight optimization was performed using validation RMSE as the objective, which may not fully capture downstream clinical utility. Alternative objectives—such as event-based metrics or risk-weighted costs—could be explored in future work. Third, evaluation was conducted retrospectively on two datasets; prospective validation within real-time clinical workflows or integration into AID systems is necessary to determine real-world impact. Fourth, although the AZT1D dataset includes real-world data from 25 individuals, broader demographic and physiological diversity is required to fully assess generalizability across age groups, treatment regimens, and sensor technologies.

Future research may extend this framework in several directions. Longer prediction horizons (e.g., 120 minutes) could be explored, though increased temporal uncertainty may require additional contextual or behavioral features. Another promising direction is adaptive or patient-specific threshold optimization, jointly learning glycemic boundaries and regional weights to better align with individualized treatment targets. Finally, integrating clinically interpretable risk measures into the optimization objective may further strengthen the connection between predictive modeling and therapeutic decision support.

Overall, these results suggest that clinically-aware training objectives can meaningfully enhance glucose forecasting without increasing architectural complexity. By focusing on how models are trained rather than solely on how they are structured, GLIMMER demonstrates a complementary pathway for improving both predictive performance and clinical safety in type 1 diabetes management.

\section{Conclusion}
In this article, we introduced GLIMMER, a modular and architecture-agnostic training framework with a custom region-aware loss function designed to improve blood glucose forecasting and enhance clinical decision-making in type~1 diabetes management. Rather than proposing a specific model architecture, GLIMMER can be integrated with existing forecasting models to improve prediction quality, particularly in clinically significant dysglycemic regions where accurate forecasts are most critical. Our extensive analysis using the OhioT1DM dataset showed that GLIMMER reduces RMSE and MAE by 25.1\% and 31.2\%, respectively, when applied to a CNN-LSTM architecture, while the Transformer variant achieves reductions of 7.0\% and 10.3\%. Similar trends were observed on the AZT1D dataset, with reductions of 23.9\% in RMSE and 27.9\% in MAE for CNN-LSTM, and 9.8\% and 13.0\% for Transformer. In addition to these improvements in overall error metrics, GLIMMER enhances prediction accuracy in dysglycemic regions, supporting better detection of high-risk events. Compared to state-of-the-art models with millions of parameters, GLIMMER achieves comparable accuracy with 2--3 orders of magnitude fewer parameters, enabling deployment without high computational cost. These results demonstrate that GLIMMER improves both accuracy and clinical reliability across model families while remaining computationally lightweight and suitable for integration into mobile applications and AID systems to support proactive and personalized diabetes management.

\section{Acknowledgment}
This work was supported in part by the Mayo Clinic and Arizona State University Alliance for Health Care Collaborative Research Seed Grant Program under Award Number ARI-320598, and in part by the National Science Foundation under grant IIS-2402650, and the National Institute of Diabetes and Digestive and Kidney Diseases of the National Institutes of Health under grant T32DK137525. Any opinions, findings, conclusions, or recommendations expressed in this material are those of the authors and do not necessarily reflect the views of the funding organizations.

\section*{Declarations}

\begin{itemize}
\item \textbf{Funding:} Mayo Clinic and Arizona State University Alliance for Health Care Collaborative Research Seed Grant Program under Award Number ARI-320598, and in part by the National Science Foundation under grant IIS-2402650 and the National Institute of Diabetes and Digestive and Kidney Diseases of the National Institutes of Health under grant T32DK137525. Any opinions, findings, conclusions, or recommendations expressed in this material are those of the authors and do not necessarily reflect the views of the funding organizations.
\item \textbf{Conflict of interest/Competing interests:} N/A
\item \textbf{Ethics approval and consent to participate:} The AZT1D data collection study was approved by the Institutional Review Board of the Mayo Clinic (\#23-003065). Written informed consent was obtained from all individual participants included in the study.
\item \textbf{Consent for publication:} Written informed consent for publication was obtained from all participants
\item \textbf{Data availability:} Data available at: \url{https://data.mendeley.com/datasets/gk9m674wcx/1}
\item \textbf{Materials availability:} N/A
\item \textbf{Code availability:} Code available at: \url{https://github.com/SamanKhamesian/GLIMMER} 
\item \textbf{Author contribution:} Saman Khamesian and Hassan Ghasemzadeh wrote the manuscript. Saman Khamesian and Asiful Arefeen implemented the code and created the figures. Bithika M. Thompson contributed to the construction of the new dataset (AZT1D) by coordinating data collection from 25 individuals with Type 1 Diabetes. All authors reviewed and evaluated the manuscript.
\end{itemize}

\bibliography{main}
\end{document}